# SWRL2SPIN: A tool for transforming SWRL rule bases in OWL ontologies to object-oriented SPIN rules

Nick Bassiliades

Author Note

Nick Bassiliades, Department of Informatics, Aristotle University of Thessaloniki, GR-54124 Thessaloniki, Greece, nbassili@csd.auth.gr

Correspondence concerning this article should be addressed to:

Nick Bassiliades, Department of Informatics, Aristotle University of Thessaloniki, GR-54124 Thessaloniki, Greece. E-mail: nbassili@csd.auth.gr

**Paper title:**

# *SWRL2SPIN: A Tool for Transforming SWRL Rule Bases in OWL Ontologies to Object-Oriented SPIN Rules*

## Abstract

Semantic Web Rule Language (SWRL) combines OWL (Web Ontology Language) ontologies with Horn Logic rules of the Rule Markup Language (RuleML) family. Being supported by ontology editors, rule engines and ontology reasoners, it has become a very popular choice for developing rule-based applications on top of ontologies. However, SWRL is probably not going to become a WWW Consortium standard, prohibiting industrial acceptance. On the other hand, SPIN (SPARQL Inferencing Notation) has become a de-facto industry standard to represent SPARQL rules and constraints on Semantic Web models, building on the widespread acceptance of SPARQL (SPARQL Protocol and RDF Query Language). In this paper, we argue that the life of existing SWRL rule-based ontology applications can be prolonged by converting them to SPIN. To this end, we have developed the SWRL2SPIN tool in Prolog that transforms SWRL rules into SPIN rules, considering the object-orientation of SPIN, i.e. linking rules to the appropriate ontology classes and optimizing them, as derived by analysing the rule conditions.

# 1. Introduction

Rule-based systems have been extensively used in several applications and domains, such as e-commerce, personalization, games, businesses and academia. They offer a simplistic model for knowledge representation for both domain experts and programmers; experts usually find it easier to express knowledge in a rule-like format and programmers usually find rule-based programming easier to understand and manipulate, decoupling computation from control. The first is performed by the rules whereas the latter is determined by the rule engine itself, that is when and how to apply the rules.

The Semantic Web initiative of the World Wide Web Consortium (W3C, 2013) works on standards, technologies and tools in order to give to the information a well-defined meaning, enabling computers and people to work in better cooperation. Ontologies can be considered as a primary key towards this goal since they provide a controlled vocabulary of concepts, each with explicitly defined and machine processable semantics. The Web Ontology Language (OWL) (Hitzler et al., 2012) is the W3C recommendation for creating and sharing ontologies on the Web. It provides the means for ontology definition and specifies formal semantics on how to derive new information.

There are mainly two modeling paradigms for the Semantic Web (Horrocks et al., 2005). The first paradigm is based on the notion of the Description Logics (Baader et al., 2010) on which the OWL is based. In this case, the semantics of OWL ontologies can be handled by DL reasoning systems, such as Pellet (Sirin et al., 2007), RacerPro (Haarslev et al., 2012), Fact++ (Tsarkov & Horrocks, 2006) and HermiT (Glimm et al., 2014) that reuse existing DL algorithms, such as tableaux-based algorithms (Baader & Sattler, 2001). The other paradigm is based on Horn logic, whereas a subset of the OWL semantics is transformed into rules that are used by a rule engine in order to infer implicit knowledge. There are major differences between these two paradigms, including computational and expressiveness aspects. For example, the DL reasoning engines have a rather inefficient instance reasoning performance, whereas rules are insufficient to model certain situations related to the open nature of the Semantic Web. The selection of the most suitable modeling paradigm depends on the domain and the needs of the application.

Since description logics and Horn logic are orthogonal in the sense that neither of them is a subset of the other (Grosof et al., 2003), there are two interesting combinations of ontologies and rules, namely their intersection, which is OWL 2 RL, and their union, namely SWRL. OWL 2 RL (Motik et al., 2012) is an OWL 2 profile is aiming at applications that require scalable reasoning without sacrificing too much expressive power. This is achieved by defining a syntactic subset of OWL 2 which is amenable to implementation using rule-based technologies, namely it is the largest syntactic fragment of OWL2 DL that is implementable using rules. The design of OWL 2 RL was inspired by Description Logic Programs (Grosof et al., 2003) and pD* (ter Horst, 2005). Obviously, OWL 2 RL is a decidable language, but one that is necessarily less expressive than either the description logic or rules language from which it is formed.

SWRL (Horrocks et al., 2004; 2005) is a semantic web rule language that combines OWL ontologies with Horn Logic rules of the RuleML family of rule languages (“RuleML”, n.d.), extending the set of OWL axioms to include Horn-like rules. SWRL is considerably more powerful than either OWL DL or Horn rules alone; however, key inference problems for SWRL are undecidable (Horrocks et al., 2005). Decidability can be regained by restricting the form of admissible rules, by imposing a suitable safety condition (Motik et al., 2005).

Being supported by the Protégé ontology editor (“Protégé”, n.d.) as well as by popular rule engines and ontology reasoners, such as Jess (Friedman-Hill, 2003), Drools (“Drools”, n.d.)

and Pellet (Sirin et al., 2007), SWRL has become a very popular choice for developing rule-based applications on top of ontologies (Billet et al., 2011; Dautov et al., 2017; Herrero-Zazo et al., 2015; Khan et al., 2017; Matheus et al., 2005; Namahoot et al., 2016; O'Connor et al., 2008; Somodevilla et al., 2015). However, SWRL being around for more than 10 years now, it is most probable that it will never become a W3C standard; therefore, its scope is difficult to reach out to the industrial world.

On the other hand, SPIN (Knublauch et al., 2011) has become a de-facto industry standard to represent SPARQL rules and constraints on Semantic Web models, building on the wide-spread acceptance of the SPARQL query language (Harris & Seaborne, 2013) for querying and processing Linked Open Data. SPARQL is well supported by numerous engines and databases. This means that SPIN rules can be directly executed on the databases and no intermediate engines with communication overhead need to be introduced. Also, SPIN is more expressive than SWRL, because SPARQL has various features such as *UNION*, *OPTIONAL*, *FILTER* and *NOT EXISTS* expressions. SPIN has an object-oriented model that arguably leads to better maintainable models than SWRL's flat rule lists. Finally, SPIN goes far beyond being just a rule language, and also provides means to express constraints and to define new functions and templates.

Furthermore, recent industrial rule-based applications (Fortineau et al., 2014; Aarnio et al., 2016; Samavi & Consens, 2018) have identified some SWRL limitations for modelling application business rules, such as the Open World Assumption and the difficulties to manage rule complexity and information update, proposing the use of SPARQL/SPIN as a rule language for OWL based models, to overcome the above issues.

For all the above reasons, in this paper, we argue that the life and expressiveness of existing SWRL rule-based ontology applications can be extended by being transformed into SPIN. To this end, we have developed a tool called SWRL2SPIN, using SWI-Prolog (Wielemaker et al., 2012) that takes as input an OWL ontology that contains an SWRL rule base and transforms SWRL rules into SPIN rules in the same ontology, taking into consideration the object-oriented scent of SPIN, i.e. linking rules to the appropriate ontology classes as derived by analyzing the rule conditions. Furthermore, conditions of transformed rules are optimized according to the hosting class by re-ordering condition elements. Our SWRL2SPIN tool is accompanied by a rich implementation of SWRL built-ins (41); however, the way these built-ins have been translated provides room for extensibility in the future to increase coverage.

So, concluding, the main contribution of the paper is that it introduces a tool for automatically transforming and optimizing flat SWRL rules to object-oriented SPIN rules, and its novelty lies on the fact that, to the best of our knowledge, there is no other tool in the literature that does that.

In the rest of the paper, we briefly review related works on SWRL rule transformations for interchange and/or execution reasons in section 2, and then, we overview SWRL and SPIN syntax and semantics, focusing on their RDF vocabularies, in sections 3 and 4, respectively. In section 5 we present our tool, its transformation methodology, how rules are embedded into classes, how they are optimized and how built-ins have been implemented. In section 6 we evaluate the tool and finally, in section 7, we conclude.

## 2. Related Work

To the best of our knowledge there is no other tool for transforming SWRL rule bases to SPIN rules. In this section, we briefly review existing approaches to transforming SWRL rules into another rule formalism, mainly for execution reasons, i.e. to be able to implement an SWRL rule engine by re-using another rule engine. The purpose of our work is rather to transform

SWRL rules into SPIN rules so that SPIN-compliant ontology / rule editors can be used to maintain / extend these rule bases / applications. Thus, it is not exactly similar to just running SWRL rules, but it could also be viewed in this way, since TopSPIN (TopQuadrant, n.d.) is a rule engine that after the translation of SWRL rules into SPIN rules it is able to execute them and store the results within the OWL ontology.

DL-reasoners do not support the full specification of SWRL because the reasoning becomes undecidable. So, there are different approaches of combining OWL-DL with SWRL reasoning.

*Translate SWRL into First Order Logic* and demonstrate reasoning tasks with a theorem prover. The best-known implementation of this paradigm is Hoolet (Bechhofer, 2004), which is an implementation of an OWL-DL reasoner that uses the first order theorem prover Vampire (Riazanov & Voronkov, 2002) and supports SWRL (Horrocks et al., 2005).

*Translate OWL-DL axioms into rules* and give the axiom rules plus the translated SWRL rules to a rule engine. This approach cannot cover the full expressivity of OWL-DL due to many incompatibilities between Description Logics and Horn Logic. Examples of this paradigm are the implementations of the SWRLTab (O'Connor et al., 2005) of Protégé ("Protégé", n.d.), such as the older SWRLJessTab plugin (Golbreich, 2004), which is available only for the v. 3.5 of Protégé and the newer SWRLDroolsTab plugin ("SWRL Drools Tab", 2012), which is available both for Protégé 5.2 and 3.5. These plugins translate the ontology axioms into facts of the Jess (Friedman-Hill, 2003) and Drools ("Drools", n.d.) production rule engines, respectively, and SWRL rules into production rules that materialize the conclusions of the rules as derived facts. Then the production rule engines run the SWRL rules along with the entailment rules that implement part of the DL-reasoning process and the results (materialized inferred axioms of the SWRL rules) are copied back to Protégé.

Other systems that support a similar functionality, using forward chaining rule engines are Bossam (Jang & Sohn, 2004)., a forward chaining deductive (Datalog) rule engine that supports SWRL with minimal built-in support for math and string functions, and SWRL2COOL (Rigas et al., 2012) a translator of SWRL rules into CLIPS production rules to accompany the O-Device production rule-based OWL reasoner (Meditskos & Bassiliades, 2008), with also limited support for math and comparison built-ins.

There are also approaches (Hirankitti & Xuan, 2011; Samuel et al., 2008) to integrate OWL axioms and SWRL rules into backward chaining rule engines, such as Prolog, which also have limited built-in support. Furthermore, the SWRL-IQ plugin (Elenius, 2012) for Protégé 3.x supported backward chaining querying of OWL ontologies and SWRL rule reasoning based on XSB Prolog. SWRL-IQ supports 47 SWRL built-ins, including *List* built-ins.

*Expand an existing OWL-DL reasoner* based on the tableaux algorithm. Most popular DL-reasoners, such as Pellet (Sirin et al., 2007), HermiT (Glimm et al., 2014), Racer (Haarslev et al., 2012), do support SWRL reasoning for DL-safe rules. Pellet supports all the SWRL built-ins except for *Lists* and provides support for only the first 5 Built-ins for *Date, Time, and Duration*. Pellet will almost certainly never support the *List* built-ins because of OWL DL restrictions. However, all *Date, Time, and Duration* built-ins could be provided in the future. HermiT also supports SWRL DL-safe rules, but with no built-in support (Glimm et al., 2009). Racer (Haarslev et al., 2012) supports processing of rules in a SWRL-based syntax by translating them into nRQL rules; there is no evidence as to whether SWRL built-ins are supported. Protégé 5.2 ("Protégé", n.d.) includes a Rules view in its Ontology Views that supports SWRL rules through the above DL reasoners.

KAON2 (Motik, n.d.), a rather different DL-reasoner based on reducing a SHIQ(D) knowledge base to a disjunctive Datalog program also supports the so-called DL-safe subset (Motik et al., 2005) of SWRL, but again without any evidence for built-in support. KAON2 can be integrated to Protégé through a DIG interface (Bechhofer, 2003).

Stardog (Stardog Union, 2018) is an RDF database or triplestore that rewrites queries to answer questions using SWRL inferences. Stardog supports two different syntaxes for defining rules. The first is native Stardog Rules syntax based on SPARQL. The second is the de facto standard RDF/XML syntax for SWRL. It has the advantage of being supported in many tools; but it's syntax is awkward. Stardog supports 55 SWRL built-ins.

Finally, there are some works (Milanovic et al., 2009; Wang et al., 2010) that translate SWRL rules into a rule meta-model for rule interchange reasons among various rules formats, such as R2ML (REWERSE Rule Markup Language) (Milanovic et al., 2009; REWERSE, 2006) and RIF (Kifer & Boley, 2013; Wang et al., 2010). The purpose of these translations is merely the interchange of rules, preserving the basic SWRL atom semantics, but without covering the built-ins which are left to the implementer.

In a completely opposite setting, (Woensel et al., 2014) implemented a cross-platform performance benchmark framework for mobile reasoning engines that supplies 1) a generic, standards-based Semantic Web layer on top of existing mobile reasoning engines; and 2) a benchmark engine to investigate and compare mobile reasoning performance. The above platform supports the SPIN rule language as a uniform rule language and RDF as a uniform data model and language. The supplied benchmark rulesets and datasets are converted by the platform to the custom rule and data formats of the various supported reasoning engines, namely AndroJena ("Androjena", n.d.), RDFQuery ("RDFQuery", n.d.), RDFStore-JS ("RDFStore-JS", n.d.) and Nools ("Nools", n.d.). Therefore, the purpose of (Woensel et al., 2014) is to translate SPIN rules to other rule formats, whereas in our case SPIN is the target language of the translation. However, our SWRL2SPIN tool could play the role of the bridge between existing SWRL rule bases that need to be deployed on one of the reasoning engines supported by (Woensel et al., 2014). In this case, SWRL2SPIN would translate SWRL rules to SPIN rules and then the Mobile Benchmark Framework of (Woensel et al., 2014) would translate SPIN rules into the corresponding mobile reasoning engine rule format.

## 3. Semantic Web Rule Language

The Semantic Web Rule Language (SWRL) (Horrocks et al., 2004) is a proposed language for the Semantic Web that can be used to express rules, combining OWL DL or OWL Lite with the Unary/Binary Datalog RuleML sublanguages of the Rule Markup Language. SWRL extends the set of OWL axioms to include Horn-like rules. It thus enables Horn-like rules to be combined with an OWL knowledge base. SWRL has the full power of OWL DL, but at the price of decidability and practical implementations. However, decidability can be regained by restricting the form of admissible rules, typically by imposing a suitable safety condition (Motik et al., 2005).

Rules are of the form of an implication between an antecedent (body) and consequent (head). The intended meaning can be read as: whenever the conditions specified in the antecedent hold, then the conditions specified in the consequent must also hold. Both the antecedent (body) and consequent (head) consist of zero or more atoms. An empty antecedent is treated as trivially true (i.e. satisfied by every interpretation), so the consequent must also be satisfied by every interpretation; an empty consequent is treated as trivially false (i.e., not satisfied by any interpretation), so the antecedent must also not be satisfied by any interpretation. Multiple atoms are treated as a conjunction. Note that rules with conjunctive consequents could easily be transformed (via the Lloyd-Topor transformations (Lloyd, 1987)) into multiple rules each with an atomic consequent. Atoms in these rules can be of the form $C(x)$, $P(x,y)$, *sameAs*($x$,$y$) or *differentFrom*($x$,$y$), where $C$ is an OWL class description, $P$ is an OWL property, and $x$, $y$ are either variables, OWL individuals or OWL data values.

SWRL has various representation syntaxes: abstract, human readable, XML concrete and RDF concrete. Listing 1 shows an SWRL rule example in human readable syntax that states "when a *student* ?*s* attends a *course* ?*c* that is taught by a *professor* ?*f*, then the *student* ?*s* knows the *professor* ?*f*". This SWRL rule extends the ontology shown in Figure 1, consisting of classes *Student* and *Professor* (subclasses of *Person*) and *Course*. Notice that we will use this rule throughout the paper as a running example for the transformation procedure of SWRL2SPIN.

```
uni:Student(?s) ∧ uni:attends(?s,?c) ∧ uni:isTaughtBy(?c,?f) →
                                                    uni:knows(?s,?f)
```

Listing 1. Sample SWRL rule in human readable syntax

```
uni:s a swrl:Variable .
uni:c a swrl:Variable .
uni:f a swrI:variable .
[ rdf:type swrl:Imp ;
    swrl:body [ a swrl:AtomList ;
      rdf:first [
        a swrl:ClassAtom ;
        swrl:classPredicate uni:Student ;
        swrl:argument1 uni:2
      ] ;
      rdf:rest [
        a swrl:AtomList ;
        rdf:first [
          a swrl:IndividualPropertyAtom ;
          swrl:propertyPredicate uni:attends ;
          swrl:argument1 uni:s ;
          swrl:argument2 uni:c
        ] ;
        rdf:rest [
          a swrl:AtomList ;
          rdf:first [
            a swrl:IndividualPropertyAtom ;
            swrl:propertyPredicate uni:isTaughtBy ;
            swrl:argument1 uni:c ;
            swrl:argument2 uni:f
          ] ;
          rdf:rest rdf:nil
        ]
      ]
    ] ;
    swrl:head [
      a swrl:AtomList ;
      rdf:first [
        a swrl:IndividualPropertyAtom ;
        swrl:propertyPredicate uni:knows ;
        swrl:argument1 uni:s ;
        swrl:argument2 uni:f
      ] ;
      rdf:rest rdf:nil
    ]
] .
```

Listing 2. Sample rule in SWRL RDF concrete syntax

Listing 2 shows how this rule is represented in the RDF concrete syntax. Rules are instances of the *swrl:Imp* class. The head and body of the rule are lists of atoms (*swrl:AtomList*); each atom can be one of *classAtom*, *IndividualPropertyAtom*, *DatavaluedPropertyAtom*, *SameIndividualAtom*, *DifferentIndividualsAtom*, or *BuiltinAtom*. All but the built-in atoms have one or two arguments (properties *swrl:argumentNN*); additionally, *classAtom* has a *classPredicate* property, whereas the *PropertyAtoms* have a *propertyPredicate* property. The *BuiltinAtom* construct has a list of arguments instead and the name of the built-in function. Arguments can be variables, declared as instances of the *swrl:Variable* class, datatype constants, in the *Value^^Datatype* format, or individuals, i.e. instances of an OWL class.

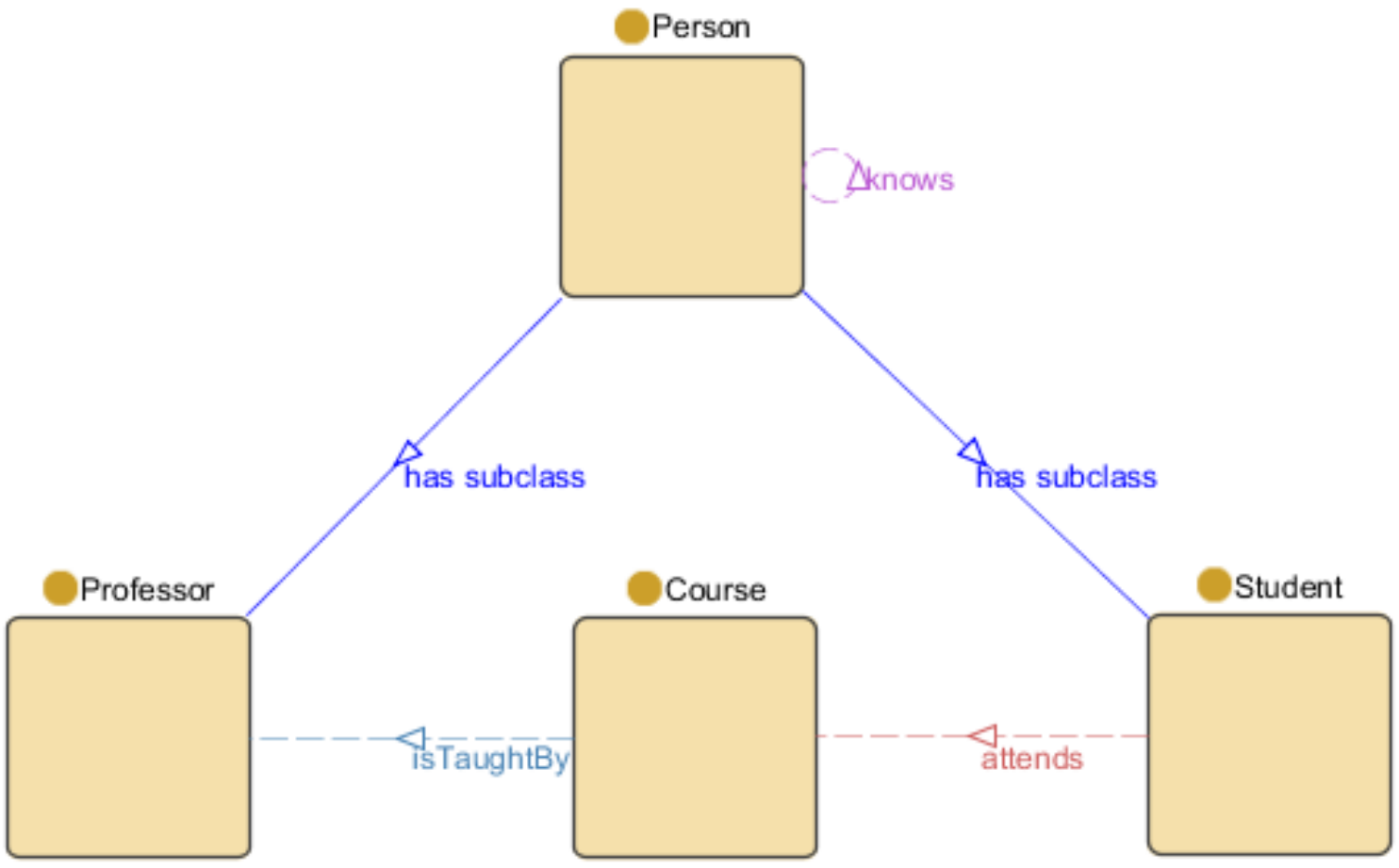

Figure 1. The ontology used in the example of Listing 1.

## 4. SPARQL Inferencing Notation

Modeling languages for the semantic web, such as RDF Schema (Brickley & Guha, 2014) and OWL (Hitzler et al., 2012), provide mechanisms for capturing the static structure of data, i.e. they are used to define classes, properties and relationships between these conceptual entities. While they define axiomatic definitions of data structures, describing general computational behavior of objects is not within their scope. On the other hand, object oriented languages provide well-known mechanisms for defining object behavior by describing classes and associating methods with class members. Object oriented methods often formalize how the modification of one attribute implies changes to other attributes. Another common purpose of methods is to capture constraints to ensure that the state of the objects remains within the bounds that the class designer had intended.

The SPARQL Inferencing Notation (SPIN) (Knublauch et al., 2011) combines concepts from object oriented languages, query languages, and rule-based systems to describe object behavior on the semantic web. One of the basic ideas of SPIN is to link class definitions with SPARQL queries to capture constraints and rules that formalize the expected behavior of those classes. SPARQL is used because it is an existing WC3 standard (Harris & Seaborne, 2013) with well-formed query semantics across RDF data, has existing widespread use amongst most RDF query engines and graph stores, and provides sufficient expressivity for both queries and general computation of data. To facilitate storage and maintenance, SPARQL queries are represented in RDF triples, using the SPIN SPARQL Syntax (Knublauch, 2011b).

The SPIN Modeling Vocabulary (Knublauch, 2011a) defines a collection of properties and classes that can be used to link RDFS and OWL classes with SPARQL queries. For example, the class *ex:Department* can define a property *spin:rule* that points to a SPARQL *CONSTRUCT* query that computes the value of *ex:studentProfessorRatio* based on the values of *ex:enrolledStudents* and *ex:numberOfFaculty*. These properties follow existing SPARQL standards, and the execution of these constructs can be efficiently handled by any SPARQL processor. Since SPIN is entirely represented in RDF, rules and constraints can be shared on the web together with the class definitions they are associated with. The attachment of rules to classes also encourages a style in which rules are locally scoped and thus easier to maintain, avoiding the spaghetti code of "flat" rule languages, such as SWRL.

Other important features of SPIN include (a) SPIN templates, which are parameterized SPARQL queries, and (b) SPIN functions, i.e. user-defined SPARQL functions. Both are not further discussed in the scope of this paper, since SWRL2SPIN does not use them.

The SPIN class description vocabulary defines several RDF properties that can be used to attach SPARQL queries to classes. The property *spin:rule* can be used by SPIN reasoning engines to construct inferred RDF triples from the currently asserted information in the model. The SPARQL queries referenced by the SPIN properties are interpreted in the context of the associated class. At run-time, the SPARQL variable ?*this* is (by default) pre-bound with instances of the class and its sub-classes. Typically, the query itself does not need to bind ?*this* to any value in the *WHERE* clause. The execution context (e.g., inference engine) will do this before the query is executed.

SPIN takes an object-oriented world view on Semantic Web models, in which SPARQL queries play a similar role to functions and methods. Inheritance (expressed using *rdfs:subClassOf*) is treated in the sense that any query/rule defined for super-classes will also be applied to subclasses. In other words, SPIN class descriptors can only "narrow down" and further restrict what has been defined further up in the class hierarchy. In this spirit, global class descriptions are those that are attached to the root class *rdfs:Resource* or its OWL equivalent *owl:Thing*. Those global queries may not even mention ?*this* at all.

The property *spin:rule* links an *rdfs:Class* with a SPARQL *CONSTRUCT* query that defines an inference rule that determines how additional triples can be inferred from what is stated in the *WHERE* clause. For each binding of the pattern in the *WHERE* clause of the rule, the triple templates from the *CONSTRUCT* clause are instantiated and added as inferred triples to the underlying model. At query execution time, the SPARQL variable ?*this* is bound to the current instance of the class.

The example in Listing 3 defines a SPIN rule (in textual SPARQL format), attached to class *uni:Student* via the *spin:rule* property, that infers the value of the *uni:knows* property from values of *uni:attends* and *uni:isTaughtBy*. Listing 4 shows how the same rule is represented using the SPIN modeling vocabulary. Notice that this SPIN rule is equivalent to the SWRL rule at Listing 1 and it is the result of the SWRL2SPIN translation procedure that will be presented in Section 5.

```
uni:Student
  a rdfs:Class ;
  spin:rule
    [ a sp:Construct ;
      sp: text """
        CONSTRUCT {
          ?this uni:knows ?faculty .
        }
        WHERE {
        ?this uni:attends ?course .
        ?course uni:isTaughtBy ?faculty
        }"""
    ].
```

Listing 3. Sample SPIN rule

```
[ a   sp:Construct ;
  sp:templates ([ sp:object spin:_this;
                  sp:predicate uni:knows ;
                  sp:subject sp:_faculty
                ]) ;
```

```
  sp:where ([ sp:object spin:_this ;
              sp:predicate uni:attends ;
              sp:subject sp:_course
            ]
            [ sp:object sp:_course;
              sp:predicate uni:knows ;
              sp:subject sp:_faculty
            ])
]
```

Listing 4. Sample rule using the SPIN modeling vocabulary

SPIN rules are instances of the *sp:Construct* class; the rule “head” is defined with the *sp:templates* property whereas the *sp:where* property defines the rule “body”. The above properties contain lists of triple patterns (*sp:subject*, *sp:predicate*, *sp:object*). Other SPARQL query elements contained in rule “body” can be *TriplePath*, *Filter*, *Bind*, *SubQuery*, *Optional*, *Union*, *NamedGraph*, *NotExists*, *Minus*, *Service*, and *Values*. In the following we only present in detail the first four, since they are the only ones used in the SWRL2SPIN tool.

A *TriplePath* is similar to a triple pattern, but instead of an *sp:predicate*, has an *sp:path* property, whose value can be one of several types, *sp:SeqPath* being the most usual one. The sequential steps of the path are represented through consecutive *sp:pathNN* properties. The representation is more complex when arbitrary length path matching is involved, i.e. when the * operator is used.

Filter elements are blank nodes, instances of *sp:Filter* that have property *sp:expression*, pointing to an expression that can be evaluated to true or false. Expressions are actually function calls which are represented as instances of the function's URI. All other properties of expressions (or function calls) are interpreted as arguments, using consecutive *sp:argNN* properties. However, other property names can be used as well, depending in the function. Arguments can be either datatype constants or variables, which are blank nodes with an *sp:varName* property whose value is a string. E.g. the `FILTER (?y > 30)` expression is shown in Listing 5.

The *BIND* keyword assigns a computed value to a variable. Bind assignments in the rule “body” are represented as instances of the class *sp:Bind*, having an *sp:variable* property to point at the variable on the right side of the assignment. The property *sp:expression* points to the root of the expression tree that delivers the computed value, in much a similar way to filter expressions (i.e. function calls). E.g., the expression `BIND ((?x * 2) AS ?y)` is shown in Listing 6.

```
[ rdf:type sp:Filter ;
  sp:expression [  rdf:type sp:gt ;
      sp:arg1 [
          sp:varName "y"^^xsd:string ;
        ] ;
      sp:arg2 "30"^^xsd:int ;
    ] ;
]
```

Listing 5. Filter expression in SPIN modeling vocabulary

```
[ rdf:type sp:Bind ;
  sp:expression [  rdf:type sp:mul ;
      sp:arg1 [ sp:varName "x" ; ] ;
      sp:arg2 2 ;
    ] ;
  sp:variable [ sp:varName "y" ; ] ;
]
```

Listing 6. Bind expression in SPIN modeling vocabulary

The *SubQuery* element is used to embed SPARQL queries within other queries to achieve results, such as limiting the number of results from some sub-expression within the query. Although conditions are flat in SWRL and no embedding is possible, in SWRL2SPIN the *SubQuery* element is used to translate the *swrlb:length* built-in (see Section 5.3). Sub-queries are represented as blank instances of the class *sp:SubQuery*. The property *sp:query* points to the nested query, which is a blank instance of the appropriate SPARQL query type, e.g. *sp:Select*, with properties *sp:resultVariables* for representing the projection variables and *sp:where* for the body of the query.

The rest of the SPARQL query elements were not needed for the translation, at least as far as the current implementation of SWRL built-in functions concerns. A brief explanation for each element is given below. The *Optional* element is used for optional triple patterns in the graph, whereas all SWRL atoms are mandatory. The *Union* element would represent disjunction in the rule body; however, SWRL does not allow disjunction. The *NamedGraph* element is used to query graphs other than the default, whereas SWRL does not support such an option. The *NotExists* and *Minus* elements are used to support negation in SPARQL, whereas SWRL is a monotonic rule language with no negation support at all. The *Service* element is used to instruct a federated query processor to invoke a portion of a SPARQL query against a remote SPARQL endpoint. Such distributed rule execution features are not possible in SWRL. Finally, the *Values* element is used to assign variables with pre-specified constant values in a SPARQL query. Such a feature is not present in SWRL, where variables can only take values from matching ontology terms.

Concluding, compared to SWRL, SPIN offers the following advantages

- It is based on SPARQL, a well-established query language and protocol, which is well-supported by numerous engines and databases. This means that SPIN rules can be directly executed on the databases and no intermediate engines with communication overhead need to be introduced.
- It is more expressive, because SPARQL has various features such as UNION and FILTER expressions.
- It has an object-oriented model that leads to better maintainable models and faster rule execution than SWRL's flat rule lists. Specifically, the SPIN rule engine does not have to check all rules at all times, as it is the case for SWRL, but instead rules are checked incrementally when new instances of a certain class are inserted (or modified) in the ontology. This leads to better rule execution performance.
- It provides means to express constraints and to define new functions and templates, besides being a mere rule language.

A more detailed presentation of the SPIN modelling vocabulary and syntax can be found at the respective references and are out of the scope of this paper.

## 5. SWRL2SPIN

The SWRL2SPIN tool accepts at its input an OWL ontology with SWRL rules embedded in the ontology using the RDF concrete syntax of SWRL, as exported by tools such as Protégé combined with the SWRLtab plugin. The tool produces at its output an OWL ontology (just copying the input one) extended by SPIN rules that have been created by translating the SWRL rules. SPIN rules are embedded inside their corresponding classes, following the OO nature of SPIN, instead of having a flat rule base as in SWRL. Furthermore, the ?*this* variable of SPIN is used to identify instances of the rule-embedding class, therefore SWRL condition elements that identify the class of the corresponding instances are removed, speeding-up, thus, rule execution. Finally, the same SWRL may involve instances of multiple classes, so our tool generates multiple versions / views of a rule, optimized for each of the classes, separately.

Formally, in SWRL2SPIN the input is an ontology $O_{inp} = < C, P, I, R_{swrl} >$, where $C$ is the set of classes of the ontology, $P$ is the set of properties, $I$ is the set of instances, and $R_{swrl}$ is the set of SWRL rules, and the output is the ontology $O_{out} = < C', P, I, R_{spin} >$, where $C'$ is the set of the modified classes of the input ontology and $R_{spin}$ is the set of SPIN rules that have been translated from SWRL rules and linked from within the classes in $C'$. Therefore, it holds that:

$$C' = \{< c, \{r_1^c, r_2^c, \ldots, r_n^c\} > | c \in C \wedge (\forall i, r_i^c \in R_{spin})$$

The above means that a class maybe linked to several SPIN rules according to the translation procedure that we will present below. The translation procedure is a mapping from a single SWRL rule to one or more SPIN rules:

$$TrnslRule: R_{swrl} \rightarrow \wp(R_{spin})$$

The main procedure for translating a SWRL rule into a SPIN rule involves mapping classes and properties of the RDF concrete syntax of SWRL into corresponding classes and properties of the SPIN modeling vocabulary, in a recursive way starting from *swrl:Imp* instances, following an almost one-to-one mapping scheme shown in Table 1. The only exception to the straightforward mapping is the SWRL built-ins whose translation is customized for each function. We will discuss translation of built-ins in section 5.3.

Table 1. Correspondence between SWRL and SPIN constructs

| SWRL | SPIN |
|---|---|
| swrl:Imp | sp:Construct |
| swrl:head | sp:templates |
| swrl:body | sp:where |
| swrl:ClassAtom<br>swrl:classPredicate <Class><br>swrl:argument1 <Arg> | sp:subject <Arg><br>sp:predicate rdf:type<br>sp:object <Class> |
| swrl:IndividualPropertyAtom<br>swrl:propertyPredicate <Prop><br>swrl:argument1 <Arg1><br>swrl:argument2 <Arg2> | sp:subject <Arg1><br>sp:predicate <Prop><br>sp:object <Arg2> |
| swrl:SameIndividualAtom<br>swrl:argument1 <Arg1><br>swrl:argument2 <Arg2> | sp:subject <Arg1><br>sp:predicate owl:sameAs<br>sp:object <Arg2> |
| swrl:DifferentIndividualsAtom<br>swrl:argument1 <Arg1><br>swrl:argument2 <Arg2> | sp:subject <Arg1><br>sp:predicate owl:differentFrom<br>sp:object <Arg2> |
| swrl:DatavaluedPropertyAtom<br>swrl:propertyPredicate <Prop><br>swrl:argument1 <Arg1><br>swrl:argument2 <Arg2> | sp:subject <Arg1><br>sp:predicate <Prop><br>sp:object <Arg2> |

| SWRL | SPIN |
|---|---|
| swrl:BuiltinAtom<br>swrl:builtin <Fun><br>swrl:arguments <Args> | *Customized translation* |
| swrl:Variable <Var> | sp:varName "<Var>" |
| <Value> ^^ <DataType> | <Value> ^^ <DataType> |
| <Individual> | <Individual> |

Formally, each SWRL rule consists of a body and a head, each of which consists of one or more SWRL atoms.

$$r_{swrl} = \langle head_{swrl}, body_{swrl} \rangle$$
$$head_{swrl} = \{atom_{swrl} | atom_{swrl}.type \in AtomTypes_{swrl}\}$$
$$body_{swrl} = \{atom_{swrl} | atom_{swrl}.type \in AtomTypes_{swrl}\}$$

$$AtomTypes_{swrl} = \left\{ \begin{matrix} ClassAtom, \\ IndividualPropertyAtom, \\ SameIndividualAtom, \\ DifferentIndividualsAtom, \\ DatavaluedPropertyAtom, \\ BuiltinAtom \end{matrix} \right\}$$

On the other hand, each SPIN rule consists of a *CONSTRUCT* clause (head) and a *WHERE* clause (body). The former consists of one or more of triple patterns, where the latter includes also *TriplePath*, *Filter* and *Bind* patterns, as explained in Section 4.

$$r_{spin} = \langle construct_{spin}, where_{spin} \rangle$$
$$construct_{spin} = \{pattern_i | pattern_i \in \{TriplePtr\}\}$$
$$where_{spin} = \{pattern_i | pattern_i \in \{TriplePtr, TriplePathPtr, FilterPtr, BindPtr\}\}$$

The translation procedure (ignoring temporarily the fact that one SWRL rule can be translated to multiple SPIN rules, due to class embedding), consists of the following translation functions:

$$TrnslRule(r_{swrl}) \coloneqq \langle TrnslAtoms(head_{swrl}), TrnslAtoms(body_{swrl}) \rangle$$

$$TrnslAtoms(Atoms_{swrl}) \coloneqq \{ptr_i | ptr_i \coloneqq TrnslAtom(atom_{i,swrl}) \wedge atom_{i,swrl} \in Atoms_{swrl}\}$$

Each SWRL atom is translated into SPIN constructs according to the mappings of Table 1. For example, the *ClassAtom* is translated as follows:

$$TrnslAtom(atom|atom.type = ClassAtom) := \langle TrnslArg(atom.argument1), type, TrnslArg(atom.classPredicate) \rangle$$

where <*x*,*y*,*z*> denotes a SPARQL triple pattern and the notation *a.p* denotes the property $p$ of the RDF instance $a$.

Another example is the *IndividualPropertyAtom* which is translated as follows:

$$TrnslAtom(atom|atom.type = IndividualPropertyAtom) := \langle TrnslArg(atom.argument1), atom.propertyPredicate, TrnslArg(atom.argument2) \rangle$$

The rest of the SWRL atoms have similar translation patterns and are omitted for brevity. The only different translation pattern is that of built-in atoms, which are presented in Section 5.3.

Finally, SWRL arguments are translated to their SPARQL equivalents, i.e. variables datatype values, individuals and classes remain unchanged, whereas variables are translated by prefixing them with a question mark.

In the following, we give an example of translating a SWRL rule without built-ins to a SPIN rule. Consider our running example, the SWRL rule at Listing 1 that is translated into the SPIN rule at Listing 7. The actual translation is performed between the RDF representations of the SWRL and SPIN rules, shown in Listing 8 and Listing 9, respectively. However, the above abstract description of the translation is still valid since there is one-to-one mapping between the abstract /textual syntaxes and the RDF representations for both rule languages.

```
CONSTRUCT {
    ?x :knows ?z .
}
WHERE {
    ?x rdf:type :Student .
    ?x :attends ?y .
    ?y :isTaughtBy ?z .
}
```

Listing 7. Translation of SRWL rule of Listing 1 into SPIN

```
:x rdf:type swrl:Variable .
:y rdf:type swrl:Variable .
:z rdf:type swrl:Variable .
[ rdf:type swrl:Imp ;
  swrl:body [ rdf:type swrl:AtomList ;
             rdf:first [ rdf:type swrl:ClassAtom ;
                        swrl:classPredicate :Student ;
                        swrl:argument1 :x   ] ;
             rdf:rest [ rdf:type swrl:AtomList ;
             rdf:first [ rdf:type swrl:IndividualPropertyAtom ;
                        swrl:propertyPredicate :attends ;
                        swrl:argument1 :x ;
                        swrl:argument2 :y   ] ;
             rdf:rest [ rdf:type swrl:AtomList ;
             rdf:first [ rdf:type swrl:IndividualPropertyAtom ;
                        swrl:propertyPredicate :isTaughtBy ;
                        swrl:argument1 :y ;
                        swrl:argument2 :z   ] ;
             rdf:rest rdf:nil
          ]]] ;
  swrl:head [ rdf:type swrl:AtomList ;
             rdf:first [ rdf:type swrl:IndividualPropertyAtom ;
                        swrl:propertyPredicate :knows ;
                        swrl:argument1 :x ;
                        swrl:argument2 :z   ] ;
             rdf:rest rdf:nil
]] .
```

Listing 8. Example of an input SWRL rule in RDF concrete syntax

```
spin:rule [
    rdf:type sp:Construct ;
    sp:templates (
        [ sp:object [ sp:varName "z" ;] ;
          sp:predicate :knows ;
          sp:subject [ sp:varName "x" ; ] ; ]   ) ;
    sp:where (
        [ sp:object :Student ;
          sp:predicate rdf:type ;
          sp:subject [ sp:varName "x" ; ] ; ]
        [ sp:object [ sp:varName "y" ; ] ;
          sp:predicate :attends ;
          sp:subject [ sp:varName "x" ; ] ; ]
        [ sp:object [ sp:varName "z" ; ] ;
          sp:predicate :isTaughtBy ;
          sp:subject [ sp:varName "y" ; ] ; ]  ) ;
] ;
```

Listing 9. Example of an output SPIN rule in SPIN modelling vocabulary

## *5.1 Embedding SPIN rules in Classes*

One of the unique features of SPIN compared to SWRL is the ability to embed rules into classes and treat them in an OO way as inheritable behaviors (aka methods). By doing so, instances of the embedding class can be identified by variable ?*this*. We do this in SWRL2SPIN as follows:

1. We identify variables in the rule body that refer to class instances that play the role of the "subject" in the triple patterns by collecting all the variables in the rule body that are: arguments of a *ClassAtom* construct or first arguments of an *IndividualPropertyAtom* or a *DatavaluedPropertyAtom* construct. The rationale behind this is that subjects of triple patterns can only play the role of the "referenced object", i.e. the object that exhibits the class behavior. Formally, in step 1 we build the set *ThisVars* as follows:

$$ThisVars(body_{swrl}) := \left\{ \langle atom.argument1, atom \rangle \middle| \begin{array}{c} atom \in body_{swrl} \wedge \\ atom.type \in \left\{ \begin{array}{c} ClassAtom, \\ IndividualPropertyAtom, \\ DatavaluedPropertyAtom \end{array} \right\} \end{array} \right\}$$

2. We identify the classes these variables refer to by: (a) checking if they are arguments of a *ClassAtom* construct or (b) retrieving the domain of arguments of *IndividualPropertyAtom* or *DatavaluedPropertyAtom* constructs. Formally, we build the set *ThisClasses* as follows:

$$ThisClasses(body_{swrl}) := \{ThisClass(ThisVar) | ThisVar \in ThisVars(body_{swrl})\}$$

$$ThisClass(ThisVar) := \left\{ \begin{array}{l} ThisVar.atom.classPredicate, ThisVar.atom.type = ClassAtom \\ ThisVar.atom.propertyPredicate.domain, ThisVar.atom.type \in \left\{ \begin{array}{c} IndividualPropertyAtom, \\ DatavaluedPropertyAtom \end{array} \right\} \end{array} \right.$$

3. We generate as many rules as the number of the different classes "discovered" in step 2. By doing so, we rewrite each rule of step 3 so that: (a) corresponding variable names are replaced by *?this*, (b) *rdf:type* triple patterns that refer to *?this* are removed from the rule body,

(c) triple patterns in the rule body are re-ordered so that the order of triple patterns is optimal.
Actually, the *TrnsRule* function that has been defined above, should be re-defined as follows:

$$TrnslRule(r_{swrl}) \coloneqq \{\langle ThCl, repl(ThCl, Trnsl(head_{swrl})), emb(ThCl, Trnsl(body_{swrl}))\rangle | ThCl \in ThisClasses(body_{swrl})\}$$

The *emb* function performs steps 3(a)-3(c) discussed above:

$$emb(ThisClass, Patterns) := opt\big(rem\big(repl(ThisClass, Patterns)\big)\big)$$

Step 3(a) is implemented by the *repl* function, which replaces occurrences of variables in the *CONSTRUCT* or the *WHERE* clause of the SPIN rule, with the variable ?*this*, when the class that the rule is embedded coincides with the class of the variable:

$$repl(ThisClass, Patterns) \coloneqq \left\{ \langle s', p, o' \rangle \middle| \begin{array}{c} \langle s, p, o \rangle \in Patterns \land s' = \begin{cases} ?\,this, ThisClass(s) = ThisClass \\ s,\ otherwise \end{cases} \\ \land\, o' = \begin{cases} ?\,this, ThisClass(o) = ThisClass \\ o,\ otherwise \end{cases} \end{array} \right\}$$

Step 3(b) is implemented by the rem function which removes all triple patterns of the form <*?this,rdf:type,Class*> from the *WHERE* clause of the SPIN rule:

$$rem(Patterns) \coloneqq Patterns - \{\langle ?\,this, rdf{:}\,type, o\rangle | \langle ?\,this, rdf{:}\,type, o\rangle \in Patterns\}$$

Finally, function *opt* that implements step 3(c) is discussed in Section 5.2.

For our running example of the SWRL rule at Listing 1, the variables that belong to set *ThisVars* are:

1. variable ?*x*, due to the *Student*(?*x*) class atom
2. variable ?*y*, due to the *isTaughtBy*(?*y*,?*z*) individual property atom.

These variables belong to classes *Student* and *Course*, respectively. The former is discovered from the *Student*(?*x*) class atom, while the latter is discovered from the domain of the *isTaughtBy*(?*y*,?*z*) individual property atom. Thus, the SWRL rule is actually converted into two SPIN rules embedded at classes *Student* (Listing 10) and *Course* (Listing 11), respectively:

```
CONSTRUCT {     # @Student
      ?this  :knows ?z .
}
WHERE {
      ?this  :attends ?y .
      ?y  :isTaughtBy ?z .
}
```

Listing 10. SPIN rule embedded at class *Student*

```
CONSTRUCT {     # @Course
      ?x  :knows ?z .
}
WHERE {
      ?x rdf:type  :Student .
      ?x  :attends ?this .
      ?this  :isTaughtBy ?z .
}
```

Listing 11. SPIN rule embedded at class *Course*

The rationale for generating multiple SPIN rules from a single SWRL rule is the following. The SWRL rule example we use involves in its condition multiple objects belonging to multiple classes (*Student*, *Course* and *Professor*). A straightforward translation would create just one rule that is stored at the *owl:Thing* class and inherited by any class of the ontology. This would mean that the rule will be checked by any update in any of the classes, resulting in lower rule execution performance. Since in SPIN we can exploit the embedding of rules inside classes, we can speedup rule execution by allowing only rules that are relevant to each class to be checked. In our example, the SWRL rule could be embedded in any of the involved classes. However, if it is embedded in only one class, e.g. *Student*, then it would only be considered when new *Student* instances are created and linked to the instances of the other classes. Using our approach, the same rule is also created for class *Course*, so that when a new course is created the rule runs as well.

The rule is not created for class *Professor*, because *Professor* instances do not appear as subjects in the SWRL condition. If there was a rule for class *Professor*, it would look like the one in Listing 12, where it is evident that this rule would be very inefficient to check, since any new instance of class *Professor* should be joined with all instances of class *Course* and all instances of class *Student*. Therefore, we have decided not to create OO SPIN rules that would not bring any execution performance advantage.

```
CONSTRUCT {     # @Professor
      ?x  :knows ?this .
}
WHERE {
      ?x rdf:type  :Student .
      ?x  :attends ?y .
      ?y  :isTaughtBy ?this .
}
```

Listing 12. SPIN rule NOT created for class *Professor*

Concluding, it is more efficient to create multiple SPIN rules embedded in all associated classes so that inserting objects in all associated classes will trigger the rule, rather than keeping a single flat rule, as in SWRL, and checking the rules with multiple joins at all times. Furthermore, we refrain from creating SPIN rules for classes their objects do not stand as subjects in the triple patterns of the SPARQL query, in order to avoid expensive joins, as well.

### *5.2 Optimizing SPIN rules*

In the example of Section 5.1 at Listing 11, the body of the SPIN rule at class Course has two triple patterns that contain variable ?*this* and one triple pattern for variable ?*x* ranging over all instances of class Student, following the initial ordering of the atoms at the body of the SWRL rule. However, it is evident that this ordering leads to a very inefficient SPARQL query execution, since variable ?*x* can be instantiated with many values, whereas variable ?this instantiates each time with only one value. So, SWRL2SPIN re-orders the triple patterns in the body of converted / embedded SPIN rules (according to the step 3(c) in Section 5.1) using the following heuristics:

1. Triple patterns that contain variable ?*this* at the subject of the triple pattern are placed first;
2. Triple patterns that contain variable ?*this* at the object of the triple pattern are placed second;
3. Triple patterns that contain the properties *owl:sameAs* or *owl:differentFrom* are placed after the triple patterns that instantiate the variables of their subject and object;

4. The order of all other triple patterns remains unchanged, so as not to alter the condition sequence of the original SWRL rule too much.

Formally, the algorithm of function *opt* (see Section 5.1) is given at Listing 13. This function uses the following auxiliary functions, that implement the above heuristics.

$$opt1(Patterns) \coloneqq \{\langle ?\,this, p, o\rangle | \langle ?\,this, p, o\rangle \in Patterns\}$$

$$opt2(Patterns) \coloneqq \{\langle s, p, ?\,this\rangle | \langle s, p, ?\,this\rangle \in Patterns\}$$

$$opt3(Patterns) \coloneqq \{\langle s, p, o\rangle \big| \langle s, p, o\rangle \in Patterns \wedge p \in \{sameAs\,, differentFrom\,\}\}$$

$$opt4(Patterns) \coloneqq Patterns - opti1(Patterns) - opt2(Patterns) - opt3(Patterns)$$

Function *opt5* generates a sequence of patterns in the SPIN rule body, where triples patterns of heuristics 1, 2 and 3 are placed. The ǁ operator denotes the concatenation of two sequences.

$$opt5(Patterns) \coloneqq opti1(Patterns) \parallel opt2(Patterns) \parallel opt4(Patterns)$$

Then, the ***foreach*** loop in the body of the *opt* function iterates over all *owl:sameAs* or *owl:differentFrom* triples (heuristic 3), and places each triple in the next position in the sequence after the triple that contains the first occurrence of the subject or object of the triple, whichever comes later in the sequence. In Listing 13 we use the following auxiliary functions, whose implementation is trivial: Function ***first-pos***(*<s,p,o>*,*Patterns*) returns the position of the first occurrence of the triple *<s,p,o>* in the sequence of triple *Patterns*, function ***max***(*N1*,*N2*) returns the maximum of two numbers, and function ***insert****(<s,p,o>,Pos,Patterns)* inserts the *<s,p,o>* triple in the *Patterns* sequence at position *Pos*.

```
Function opt(Patterns)
    PatternsIn := opt5(Patterns)
    Foreach <s,p,o> ∈ opt3(Patterns)
        Pos1 := first-pos(<s,p',o'>, PatternsIn)
        Pos2 := first-pos(<s'',p'',o>, PatternsIn)
        Pos := max(Pos1, Pos2)
        PatternsIn := insert(<s,p,o>, Pos+1, PatternsIn)
    Return PatternsIn
```

Listing 13. The optimization function for the SPIN rule body

According to the above, the triple patterns of the body of the SPIN rule at class Course (Listing 11) are re-ordered as shown in Listing 14.

```
CONSTRUCT {    # @Course
      ?x  :knows ?z .
}
WHERE {
      ?this  :isTaughtBy ?z .
      ?x  :attends ?this .
      ?x rdf:type  :Student .
}
```

Listing 14. Optimized SPIN rule at class Course

Notice that using the above algorithm, we try to change the order of the original SWRL rule condition elements as little as possible. The SWRL rules have been created by people having a specific understanding of the ontology domain and the order of the rule condition elements represents the understanding of the developer about how class instances logically interrelate. So, if the translated SPIN rules have a random or a completely different order of condition elements then the same developer that will pick up the translated rule base in SPIN will have a hard time comprehending the rules he/she has developed.

### *5.3 Implementing SWRL built-ins*

The translation of the SWRL built-ins does not follow the straightforward approach of the rest of the SWRL atoms and it depends on the nature of each function and the existence of equivalent SPIN or SPARQL functions. More specifically, SWRL specification (Horrocks et al., 2004) has defined 78 built-in functions classified across the categories: *Comparisons*, *Mathematics*, *Boolean Values*, *Strings*, *Date*, *Time and Duration*, *URIs*, and *Lists*. Currently, SWRL2SPIN implements more than half of the SWRL built-ins (41), mostly in the categories: *Comparisons*, *Mathematics*, *Strings*, and *Lists*. For the *Date*, *Time and Duration* category, we implemented only the *swrlb:date* function.

Table 2 contains all the supported built-ins, the category they belong to and how their conversion to SPIN/SPARQL was achieved. As it can be observed, the conversion of the built-ins falls into ten categories: *binary filter*, *associative infix assign*, *binary infix assign*, *unary assign*, *assign function*, *filter function*, *magic property*, *complex assign*, *complex filter*, and *complex expression*. Filter-type conversions lead to SPARQL *FILTER* Boolean expressions, whereas assign-type conversions lead to *BIND* expressions. Simple mathematical comparisons and operations are treated as binary infix mathematical operations, such as >= or -. Addition and multiplication in SWRL built-ins can have an arbitrary number of arguments, so they are treated as *associative binary infix operators*. Finally, there are also simple unary operators, e.g. *minus*.

The general atom transformation function (presented at the beginning of Section 5) for the case of built-in atoms is specialized as follows:

$$TrnslAtom(atom|atom.type = BuiltinAtom) := TrnslFun(convCat(atom.builtin), atom.builtin, atom.arguments)$$

where the *convCat* function returns the conversion category (column 3) of the built-in function (column 1), as indicated in Table 2.

Below we give the definition of the *TrnslFun* function for two representative cases: *binary filter* and *binary infix assign*.

$$TrnslFun(\text{"}binary\ filter\text{"}, fun, args) \coloneqq \text{"}FILTER\ \text{("}\&TrnslArg(args[1])\&TrnslOp(fun)\&TrnslArg(args[2])\&\text{")"}$$

where & is the string concatenation operator, $a[n]$ is a notation that returns the $n$-th element of a list $a$ and *TrnslOp* is a function that returns the operator (column 5) of the built-in function (column 1), as indicated in Table 2. The function *TrnslArg* has been discussed at the beginning of Section 5.

For the *binary infix assign* conversion category, the translation function is defined similarly:

$$TrnslFun(\text{"}binary\ infix\ assign\text{"}, fun, args) \coloneqq \text{"}BIND\ \text{(("}\&TrnslArg(args[2])\&TrnslOp(fun)\&TrnslArg(args[3])\&\text{")}\ AS\ \text{"}\&TrnslArg(args[1])\&\text{")"}$$

Table 2. SWRL2SPIN support for SWRL built-ins

| **SWRL** | | **Conversion category** | **SPIN / SPARQL** | | |
|---|---|---|---|---|---|
| **Built-in Function** | **Category** | | **function** | **op** | **expression** |
| greaterThan(?x,?y) | Compare | binary filter | sp:gt | > | FILTER (?x > ?y) |
| greaterThanOrEqual(?x,?y) | Compare | binary filter | sp:ge | >= | FILTER (?x >= ?y) |
| lessThan(?x,?y) | Compare | binary filter | sp:lt | < | FILTER (?x < ?y) |
| lessThanOrEqual(?x,?y) | Compare | binary filter | sp:le | <= | FILTER (?x <= ?y) |
| equal(?x,?y) | Compare | binary filter | sp:eq | = | FILTER (?x = ?y) |
| notEqual(?x,?y) | Compare | binary filter | sp:ne | != | FILTER (?x != ?y) |
| add(?y,?$x_1$,?$x_2$,…,?$x_n$) | Math | associative infix assign | sp:add | + | BIND ((((?$x_1$ + ?$x_2$) +…) + ?$x_n$) AS ?y) |
| multiply(?y,?$x_1$,?$x_2$,…,?$x_n$) | Math | associative infix assign | sp:mul | * | BIND ((((?$x_1$ * ?$x_2$) * …) * ?$x_n$) AS ?y) |
| subtract(?z,?x,?y) | Math | binary infix assign | sp:sub | - | BIND ((?x - ?y) AS ?z) |
| divide(?z,?x,?y) | Math | binary infix assign | sp:divide | / | BIND ((?x / ?y) AS ?z) |
| unaryPlus(?y,?x) | Math | unary assign | sp:unaryPlus | + | BIND ((+?x) AS ?y) |
| unaryMinus(?y,?x) | Math | unary assign | sp:unaryMinus | - | BIND ((-?x) AS ?y) |
| abs(?y,?x) | Math | assign function | sp:abs | | BIND (abs(?x) AS ?y) |
| ceiling(?y,?x) | Math | assign function | sp:ceil | | BIND (ceil(?x) AS ?y) |
| floor(?y,?x) | Math | assign function | sp:floor | | BIND (floor(?x) AS ?y) |
| round(?y,?x) | Math | assign function | sp:round | | BIND (round(?x) AS ?y) |
| mod(?z,?x,?y) | Math | assign function | spif:mod | | BIND (spif:mod(?x, ?y) AS ?z) |
| stringConcat(?y,?$x_1$,?$x_2$,…,?$x_n$) | Strings | assign function | sp:concat | | BIND (CONCAT(?$x_1$,…, ?$x_n$) AS ?y) |
| stringLength(?y,?x) | Strings | assign function | sp:strlen | | BIND (STRLEN(?x) AS ?y) |
| upperCase(?y,?x) | Strings | assign function | sp:ucase | | BIND (UCASE(?x) AS ?y) |
| lowerCase(?y,?x) | Strings | assign function | sp:lcase | | BIND (LCASE(?x) AS ?y) |
| substringBefore(?y,?$x_1$,?$x_2$) | Strings | assign function | sp:strbefore | | BIND (STRBEFORE(?$x_1$, ?$x_2$) AS ?y) |
| substringAfter(?y,?$x_1$,?$x_2$) | Strings | assign function | sp:strafter | | BIND (STRAFTER(?$x_1$, ?$x_2$) AS ?y) |
| substring(?y,?x,?s,?l) | Strings | assign function | sp:substr | | BIND (SUBSTR(?x, ?s, ?l) AS ?y) |
| replace(?y,?str,?$s_1$,?$s_2$) | Strings | assign function | sp:replace | | BIND (REPLACE(?str, ?$s_1$, ?$s_2$) AS ?y) |
| endsWith(?x,?y) | Strings | filter function | sp:strends | | FILTER STRENDS(?x, ?y) |
| startsWith(?x,?y) | Strings | filter function | sp:strstarts | | FILTER STRSTARTS(?x, ?y) |
| contains(?x,?y) | Strings | filter function | sp:contains | | FILTER CONTAINS(?x, ?y) |
| matches(?x,?y) | Strings | filter function | sp:regex | | FILTER REGEX(?x, ?y) |
| tokenize(?x,?y,?z) | Strings | magic property | spif:split | | ?x spif:split ( ?y ?z ) . |
| integerDivide(?z,?x,?y) | Math | complex assign | | | BIND (spif:cast(?x / ?y, xsd:integer) AS ?z) |
| pow(?pow,?x,?n) | Math | complex assign | | | BIND (spif:cast((((?$x_1$ * ?$x_2$) * …) * ?$x_n$), xsd:integer) AS ?pow) . |

| SWRL | | Conversion category | SPIN / SPARQL | | |
|---|---|---|---|---|---|
| **Built-in Function** | **Category** | | **function** | **op** | **expression** |
| normalizeSpace(?y,?x) | Strings | complex assign | | | BIND (REPLACE(REPLACE(REPLACE(?x, "\\s+", " "), "^\\s+", ""), "\\s+$", "") AS ?y) |
| date(?y,?year,?month,?day) | Date, Time, Duration | complex assign | | | BIND (spif:cast(CONCAT(spif:cast(?year, xsd:string), "-", spif:cast(?month, xsd:string), "-", spif:cast(?day, xsd:string)), xsd:date) AS ?y) |
| containsIgnoreCase(?s$_1$,?s$_2$) | Strings | complex filter | | | FILTER CONTAINS(LCASE(?s$_1$),LCASE(?s$_2$)) |
| stringEqualIgnoreCase(?s$_1$,?s$_2$) | Strings | complex filter | | | FILTER (LCASE(?s$_1$) = LCASE(?s$_2$)) |
| empty(?list) | Lists | complex filter | | | FILTER (?list = rdf:nil) |
| first(?e,?list) | Lists | complex expr | | | ?list rdf:first ?e . |
| rest(?e,?list) | Lists | complex expr | | | ?list rdf:rest ?e . |
| member(?e,?list) | Lists | complex expr | | | ?list (rdf:rest)*/rdf:first ?e . |
| length(?length,?list) | Lists | complex expr | | | {<br>SELECT ?x ?list (COUNT(?e) AS ?length)<br>WHERE {<br>?list (rdf:rest)*/rdf:first ?e .<br>}<br>GROUP BY ?x ?list<br>} . *# ?x is an instance with a property whose value is the list ?list* |

The translation functions for the rest of the conversion categories will not be presented for brevity. We believe that the reader can easily follow the conversion patterns between the first and the last columns of Table 2.

Another large category is SWRL built-in functions with an exact equivalent SPIN / SPARQL function, as e.g. *round*, *replace*, and *contains*. The conversion of these functions is straightforward; as in the *FILTER* case, all arguments of the SWRL built-in become arguments of the SPIN / SPARQL function, whereas in the *BIND* case the first argument of the SWRL built-in becomes the variable to be bound in the SPIN / SPARQL *BIND* expression, whereas the rest of the arguments of the SWRL built-in become the arguments of the SPIN / SPARQL function.

As discussed in Section 4, *FILTER* and *BIND* expressions both have an *sp:expression* property that contains the mathematical or functional SPARQL expression; *BIND* also has an *sp:variable* for the assigned variable. All expressions belong to a type, which is the name of the main SPARQL function in the expression, e.g. *sp:gt*, *sp:lcase*, etc. In the case of the complex functional expressions, the outer function is the type of the *FILTER* expression, e.g. *sp:contains* in the case of the *containsIgnoreCase* SWRL built-in. The argument list of the SWRL built-in (property *swrl:arguments*) is treated as explained above, generating *sp:argNN* properties of the SPARQL expression / function. The only exception is the *spif:cast* function, whose second argument is represented by an *arg:datatype* property. The values of the *sp:argNN* properties can be SPIN variables, datatype constants, individuals or nested SPARQL functions / expressions.

As an example, consider the SWRL rule in Listing 15 which is translated in the SPIN rule at class *Person* (Listing 16). Specifically, the RDF concrete syntax for the SWRL built-in atom is shown in Listing 17, whereas the converted SPIN / SPARQL expression is shown in Listing 18.

```
Person(?x) ∧ firstName(?x, ?y) ∧ lastName(?x, ?z) ∧
swrlb:stringConcat(?a, ?y, " ", ?z) →
                                              fullName(?x, ?a)
```

Listing 15. Sample SWRL rule with built-in

```
CONSTRUCT {     # @Person
    ?this  :fullName ?a .
}
WHERE {
    ?this  :firstName ?y .
    ?this  :lastName ?z .
    BIND (CONCAT(?y, " ", ?z) AS ?a) .
}
```

Listing 16. Sample SWRL built-in translated to SPIN/SPARQL

```
[ rdf:type swrl:BuiltinAtom ;
  swrl:builtin swrlb:stringConcat ;
  swrl:arguments [ rdf:type rdf:List ;
            rdf:first  :a ;
            rdf:rest [ rdf:type rdf:List ;
            rdf:first  :y ;
            rdf:rest [ rdf:type rdf:List ;
            rdf:first " "^^xsd:string ;
```

```
            rdf:rest (  :z ) ] ] ]
] ;
```

Listing 17. RDF syntax for the SWRL built-in example

```
[  rdf:type sp:Bind ;
   sp:expression [  rdf:type sp:concat ;
           sp:arg1 [ sp:varName "y" ; ] ;
           sp:arg2 " " ;
           sp:arg3 [ sp:varName "z" ; ];];
   sp:variable [ sp:varName "a" ; ] ;
]
```

Listing 18. RDF syntax for the converted example of Listing 17

The rest of the SWRL built-ins are treated as *Complex* cases, meaning that their translation involves the combination of more than one simple functions, as discussed above. Complex cases can be filters, assignments or general SPARQL expressions (*graph patterns*) and they are treated in an ad-hoc manner. For example, the *integerDivide* built-in is translated as a *division* and a *cast* to integer, whereas the *pow* built-in is translated as repetitive *multiplication* using recursion. *List* built-ins are of special interest because their translation cannot be performed using SPIN/SPARQL functions, but can be treated using SPARQL path expressions. For example, the *member* built-in is translated into a recursive path expression combining *rdf:first* and *rdf:rest*. The translation of the *length* built-in is the most complicated one because it requires a SPARQL subquery that counts all the elements in the list, i.e. all possible iterations of the *rdf:rest* property in the *rdf:rest** recursive path.

A special case is *magic properties* which are supported by many SPARQL engines to dynamically compute values at query time. A magic property usually is implemented by a calculation function that determines bindings of the variables on the left or right side of the predicate. SPIN enables users to define such magic properties, in a very similar way as SPIN Functions, but providing greater flexibility. In contrast to BIND/FILTER functions, magic properties can return multiple values. Furthermore, any input or output variable may be unbound; it is the task of the magic property to find their potential bindings. The magic property *spif:split* is used in SWRL2SPIN to translate the *swrlb:tokenize* SWRL built-in. The first variable of the SWRL built-in generates multiple bindings. When the *spif:split* magic property is used, the subject of the “triple pattern” generates multiple alternative bindings. Magic properties are treated in an ad-hoc manner in SWRL2SPIN, since their definition and behavior does not follow a regular pattern.

The rest of the SWRL built-ins will be implemented as a future work, most probably as complex conversion cases or as user-defined magic properties. We notice here that the only other SWRL related tool supporting functions for RDF lists is the SWRL-IQ plugin (Elenius, 2012) for Protégé 3.x.

# 6. Evaluation

To evaluate SWRL2SPIN we have initially generated use cases of a University ontology with various SWRL rules in Protégé (“Protégé”, n.d.), using the SWRLTab editor (O’Connor et al., 2005), including all the SWRL-built-ins of Table 2, except for built-ins not supported by Protégé SWRLTab. Then we have used the SWRLDroolsTab (“SWRL Drools Tab”, 2012) to run SWRL rules and to identify and record all the rule inferences. Consequently, we have converted the SWRL use cases through SWRL2SPIN and we have tested the generated SPIN rules using TopSPIN in TopBraib Composer FE (TopQuadrant, n.d.) for equivalent inferences. The results were found identical for all use cases, except the ones that could not be run in SWRL-Drools.

Furthermore, we present below some performance tests we have conducted in order to evaluate the efficiency of the translation procedure and the efficiency of the generated rules. All tests were performed on a Windows 10 PC with Intel i7-4770 @ 3.40GHz, 8 GB RAM and SSD.

First, we have performed a scalability test for the translation time of SWRL2SPIN tool. Results shown in Figure 2 indicate that the translation time is linear to the number of rules, which was expected since each rule is translated individually, even if there are common variables or other constructs among SWRL rules. Rule 1 is the rule shown at Listing 1, with 3 atoms in the rule body, while Rule 2 (shown at Listing 19) has 6 atoms in the body, including one built-in atom. The translation time per rule also depends on the number of atoms and the number and type of built-ins. The average translation time per rule is about 0,72 msec for Rule 1 and 2,82 msec for Rule 2.

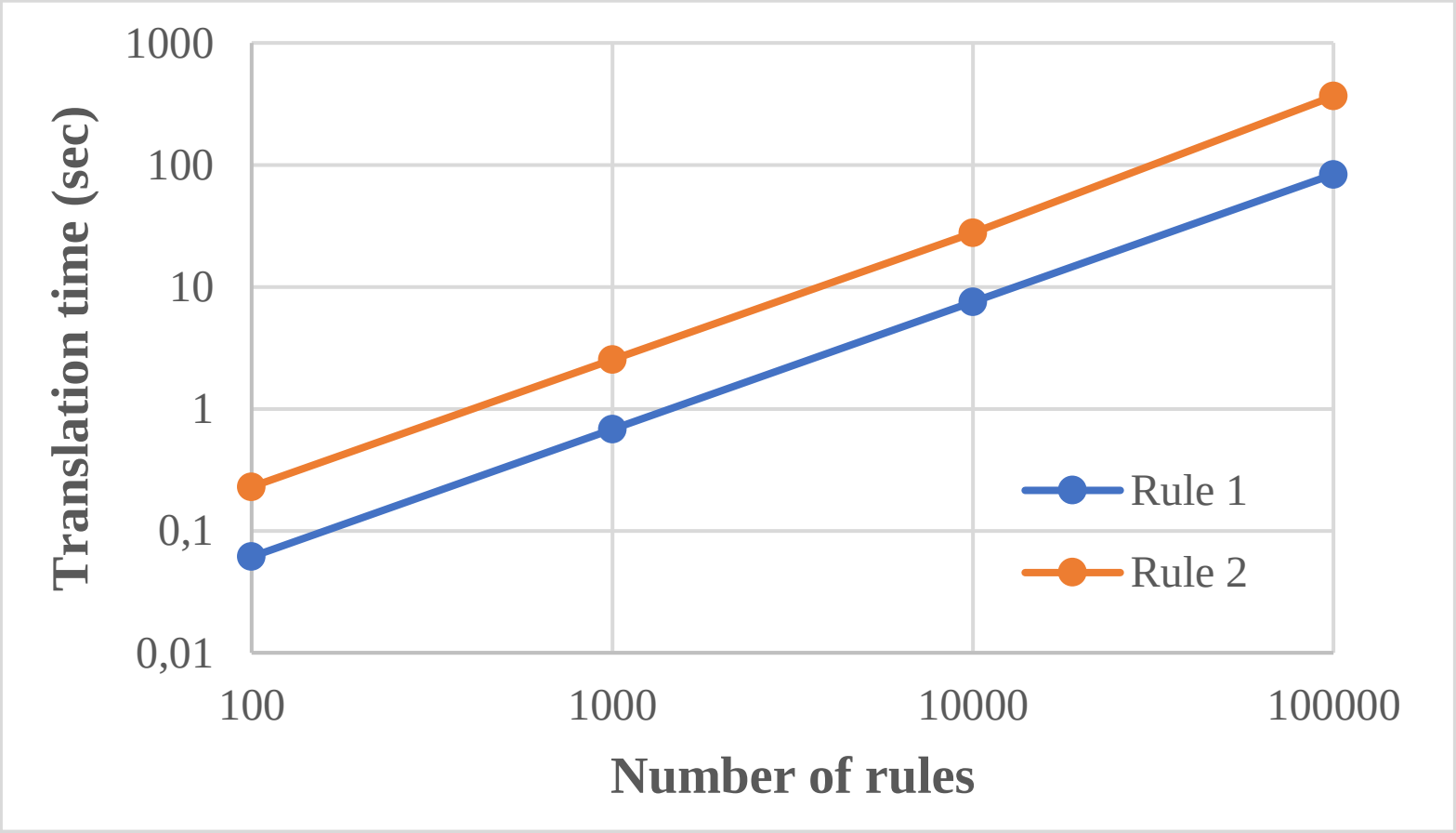

Figure 2. Rule translation time scalability.

```
Student(?x)∧attends(?x,?y)∧isTaughtBy(?y,?z) ∧ firstName(?z,?f) ∧
lastName(?z,?l) ∧ swrlb:stringConcat(?fn,?f, " ", ?l) ->
                                                    knowsName(?x, ?fn)
```

Listing 19. Sample SWRL rule in human readable syntax

The second test evaluates the performance of executing at TopBraid the SPIN rules embedded in a class against flat rules, as in SWRL. Specifically, we have tested the flat rule at Listing 7 against the rule at Listing 10 which is embedded at class *Student*. Results are shown in Table 3 and Figure 3. It is evident that rules embedded in a class perform faster than their flat equivalents. The speed improvement is 14% for an ontology with 100K *Student* instances that all attend the same course with one teacher and 32% for 1M instances and it is statistically significant with a p-value less than 0,01.

Table 3. Execution time (in msec) of a flat SPIN rule vs. a rule embedded in a class.

| | **100K instances** | **1M instances** |
|---|---|---|
| **Flat rule** | 1674 | 38098 |
| **Embedded rule** | 1438 | 25846 |
| ***Improvement*** | *14%* | *32%* |

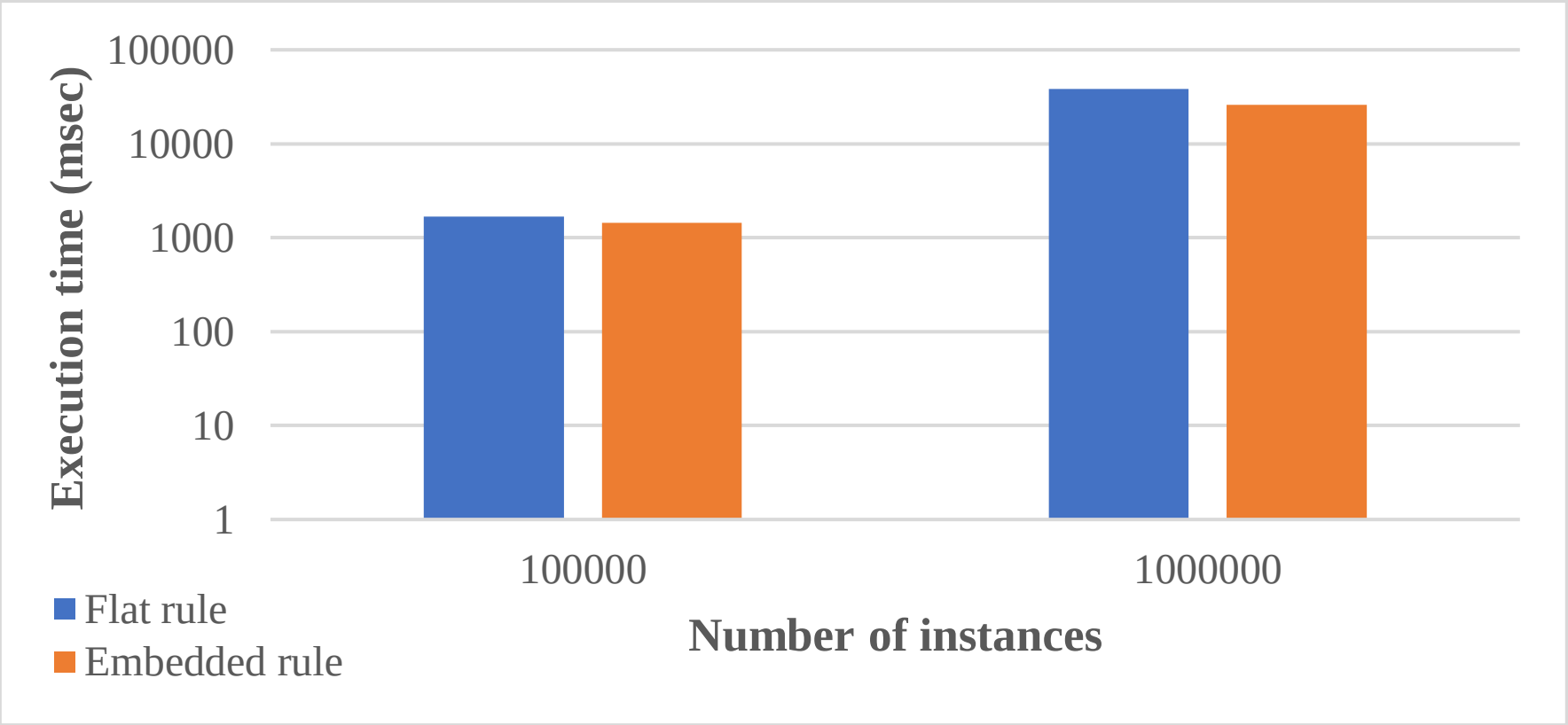

Figure 3. Performance comparison of a flat SPIN rule vs. a rule embedded in a class.

The third test tries to prove the point made at Section 5.1 that SPIN rules are embedded only at classes that their instances play the role of subjects at rule condition triple patterns and not at classes whose instances appear only as "referenced objects", i.e. they appear at triple patterns only as objects, due to performance issues. Specifically, we have tested the rule at Listing 10 which is embedded at class *Student* (we call this "*subject*" class) against the rule at Listing 12 which would have been embedded at class *Professor* (we call this "*object*" class), if we have decided to embed a rule into all references classes. Results are shown in Table 4 and Figure 4. It is evident that rules embedded in a "subject" class perform faster than rules embedded in an "object" class. The speed difference is 14% for an ontology with 100K *Student* instances and 35% for 1M instances and it is statistically significant with a p-value less than 0,01.

Table 4. Execution time (in msec) of a SPIN rule embedded in a “subject” class vs. a rule embedded in an “object” class.

| | 100K instances | 1M instances |
|---|---|---|
| **Rule embedded in “object” class** | 1677 | 39714 |
| **Rule embedded in “subject” class** | 1438 | 25846 |
| ***Improvement*** | *14%* | *35%* |

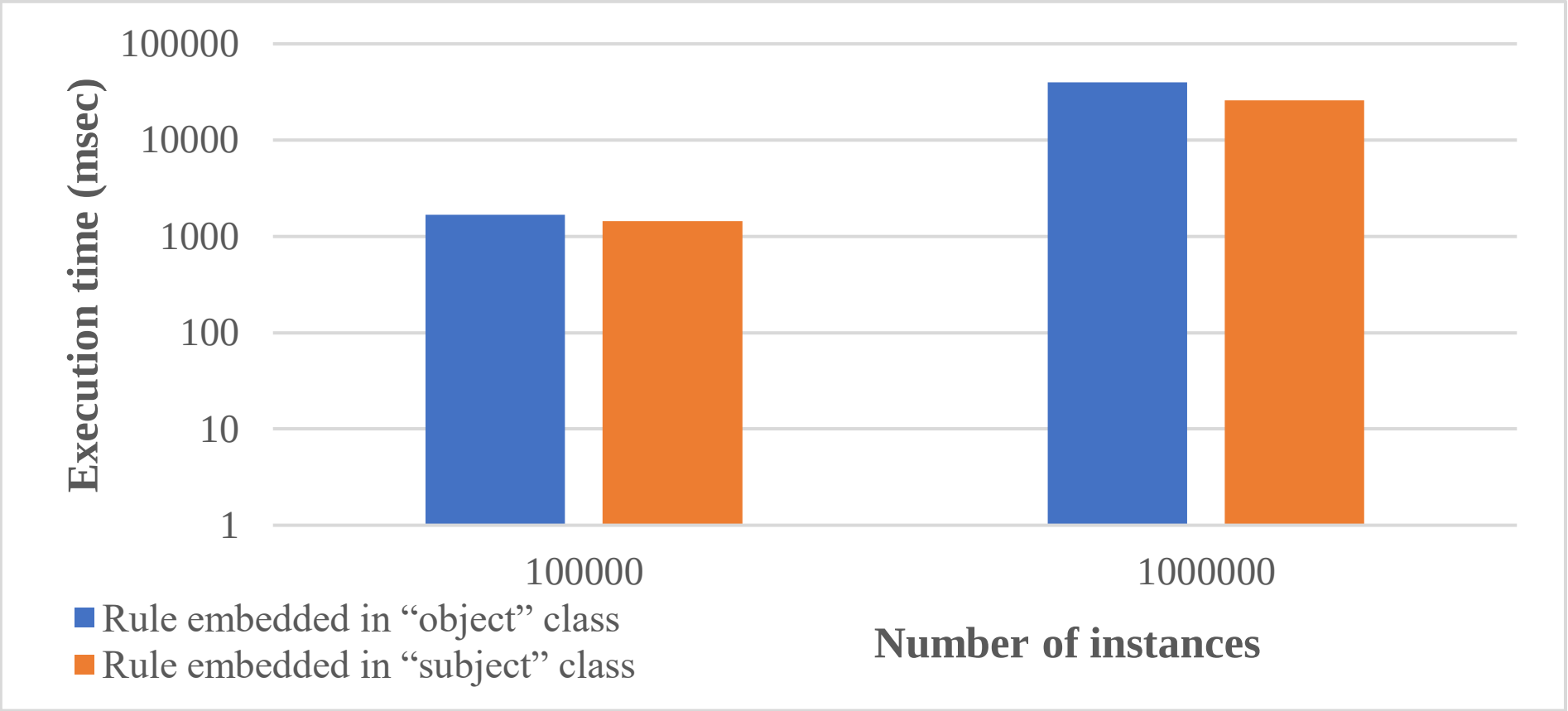

Figure 4. Performance comparison of a SPIN rule embedded in an “object” class vs. a rule embedded in a “subject” class.

Finally, the fourth test evaluates the performance of the optimized SPIN rules (section 5.2) of SWRL2SPIN against their non-optimized version. Specifically, we have tested the non-optimized rule at Listing 11 against the optimized rule at Listing 14 rule, both embedded at class *Course*. Results are shown in Table 5 and Figure 5. It is evident that optimized rules perform faster than their non-optimized equivalents. The speed improvement is 11% for an ontology with 100K instances and 30% for 1M instances and it is statistically significant with a p-value less than 0,03.

Table 5. Execution time (in msec) of an optimized SPIN rule vs. an non-optimized rule.

| | 100K instances | 1M instances |
|---|---|---|
| **Non-optimized rule** | 1413 | 29508 |
| **Optimized rule** | 1257 | 20683 |
| ***Improvement*** | *11%* | *30%* |

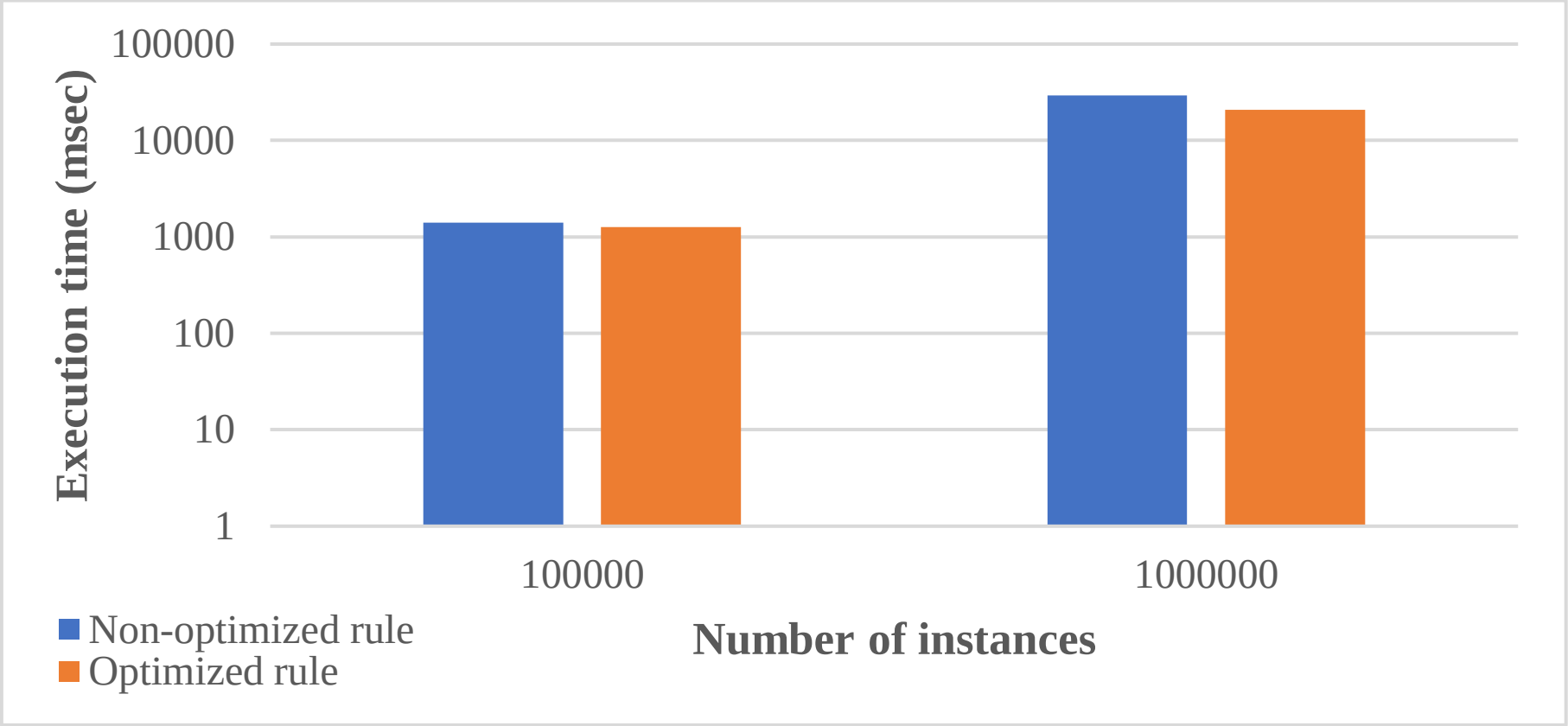

Figure 5. Performance comparison of a non-optimized SPIN rule vs. an optimized rule.

## 7. Conclusions

In this paper we have argued that SPIN is a more promising de-facto industrial standard for the future of combining ontologies and rules, because it builds upon the widespread use of SPARQL. Furthermore, SWRL has been around for quite a while, not being able to achieve a W3C recommendation status. SPIN also offers more expressivity than SWRL due to constructs like NOT EXISTS, FILTER, OPTIONAL and UNION, and also offers object-orientation by being able to store rules to classes as behaviors to be inherited through the class hierarchy. Thus, we believe that existing large SWRL projects can benefit from being translated into SPIN rules.

To this end we have developed in Prolog and presented in this paper the SWRL2SPIN prototype tool (available at: https://github.com/nbassili/SWRL2SPIN) that translates ontologies with SWRL rules into ontologies with SPIN rules. The main contribution of the paper is that the transformation translates flat SWRL rules to object-oriented SPIN rules, embedded in the appropriate ontology classes and optimized for maximum rule execution performance. The novelty of SWRL2SPIN lies on the fact that, to the best of our knowledge, there is no other tool in the literature that does that.

We have tested the tool using ontologies and SWRL rule bases edited (and tested for reasoning) by Protégé and we have successfully imported the translated ontologies and SPIN rules into the TopBraid Composer, having exactly the same inference results. We have also evaluated the scalability of the conversion tool, which is linear, and the effectiveness of the generated SPIN rules in terms of performance. Results have clearly confirmed all our claims: a) rules embedded in classes perform better than flat rules, b) rules embedded in classes that their instances play the role of "subjects" in the triple patterns of the condition perform better than rules that their instances play the role of "objects", and c) optimized embedded rules perform better than non-optimized ones. The performance improvement also depends on the size of the ontology, so for large numbers of instances our choices regarding the embedded rules perform even better.

SWRL2SPIN currently supports 41 SWRL built-ins, including built-ins for lists which are usually not supported, but we have provided a structured methodology for supporting more in the future. Notice that our translation methodology is based on direct RDF-to-RDF translation between the SWRL and SPIN RDF vocabularies; therefore, it is not dependent on the implementation language we have choose for SWRL2SPIN.

As for future work, we plan to make our tool available for public testing, to evaluate it for converting large SWRL rule bases, to support more SWRL built-ins and to be able to automate the translation process, which currently must be run from within the Prolog environment, possibly as an add-on to some SPIN rule engine. Finally, with the emergence of the SHACL language (Shapes Constraint Language) (Knublauch & Kontokostas, 2017), our future aim is to extend the tool in order to convert SWRL rules also to SHACL rules (Knublauch et al., 2017), when the latter becomes part of the SHACL recommendation. We believe that RDF validation using rule-based approaches will become really important in the future (Arndt et al., 2017) as more Linked Open Datasets become available.

## References

Aarnio, P., Vyatkin, V., & Hästbacka, D. (2016). Context modeling with situation rules for industrial maintenance. In *Proceedings of the IEEE 21st International Conference on Emerging Technologies and Factory Automation (ETFA)* (pp. 1-9). Berlin, Germany: IEEE.

Androjena. (n.d.). https://github.com/lencinhaus/androjena

Arndt, D., Meester, B.D., Dimou, A., Verborgh, R., & Mannens, E. (2017). Using Rule-Based Reasoning for RDF Validation. In *Proceedings of the International Joint Conference on Rules and Reasoning (RuleML+RR 2017)* (pp. 22-36). London, UK: Springer. Lecture Notes in Computer Science, vol. 10364.

Baader, F., Calvanese, D., McGuinness, D. L., Nardi, D., & Patel-Schneider, P. F. (Eds.). (2010). *The Description Logic Handbook: Theory, Implementation and Applications*. 2nd edition. NY, USA: Cambridge Univ Press.

Baader, F., & Sattler, U. (2001). An overview of tableau algorithms for description logics. *Studia Logica*, *69*(1), 5-40.

Bechhofer, S. (2004). Hoolet. Retrieved January 31, 2018, from http://owl.man.ac.uk/hoolet/

Bechhofer, S., (2003). The DIG Description Logic Interface: DIG/1.1, *Proceedings of DL2003 Workshop*. Rome.

Billet, Y.-G., Gravier, C., & Fayolle, J. (2011). SWRL-Based Context Awareness for Application Servers Hosting Digital Services. In *Proceedings of the 5th International Symposium on Rules: Research Based and Industry Focused (RuleML 2011 – America),* (pp. 222-229). Ft. Lauderdale, FL, USA: Springer.

Brickley, D., & Guha, R.V. (Eds.). (2014). RDF Schema 1.1, W3C Rec, 25 Feb 2014, http://www.w3.org/TR/rdf-schema/

Dautov, R., Veloudis, S., Paraskakis, I., & Distefano, S. (2017). Policy Management and Enforcement Using OWL and SWRL for the Internet of Things. In *Proceedings of the 16th International Conference on Ad Hoc Networks and Wireless (AdHoc-Now 2017)* (pp. 342-355). Messina, Italy: Springer.

Drools, (n.d.). http://www.drools.org/

Elenius, D. (2012). SWRL-IQ: A Prolog-based Query Tool for OWL and SWRL. In *Proceedings of the OWL: Experiences and Directions Workshop (OWLED 2012).* Heraklion, Greece. CEUR-WS.org, Vol. 849.

Fortineau, V., Fiorentini, X., Paviot, T., Louis-Sidney, L., & Lamouri, S. (2014). Expressing formal rules within ontology-based models using SWRL: an application to the nuclear industry. *International Journal of Product Lifecycle Management*, *7*(1), 75-93.

Friedman-Hill, E. (2003). *Jess in Action: Rule Based Systems in Java*. Manning Publications.

Glimm, B., Horridge, M., Parsia, B., & Patel-Schneider, P. F. (2009). A syntax for rules in OWL 2. In *Proceedings of the 6th International Conference on OWL: Experiences and Directions (OWLED-2009)* (pp. 29-38). Aachen, Germany: CEUR-WS.org, Vol. 529.

Glimm, B., Horrocks, I., Motik, B., Stoilos, G., & Wang, Z. (2014). HermiT: An OWL 2 Reasoner. *Journal Automated Reasoning*, *53*(3), 245-269.

Golbreich, C. (2004). Combining Rule and Ontology Reasoners for the Semantic Web. In *Proceedings of the International Workshop on Rules and Rule Markup Languages for the Semantic Web (RuleML 2004)* (pp. 6-22). Hiroshima, Japan: Springer, LNCS, Vol. 3323.

Grosof, B. N., Horrocks, I., Volz, R., & Decker, S. (2003). Description Logic Programs: Combining Logic Programs with Description Logic. In *Proceedings of the 12th International Conference on World Wide Web (WWW-2003)* (pp. 48-57). Budapest, Hungary: ACM Press.

Haarslev, V., Hidde, K., Möller, R., & Wessel, M. (2012). The RacerPro knowledge representation and reasoning system. *Semantic Web Journal*, *3*(3), 267–277.

Harris, S., & Seaborne, A. (2013). SPARQL 1.1 Query Language, W3C Rec, 21 Mar 2013. http://www.w3.org/TR/sparql11-query/

Herrero-Zazo, M., Segura-Bedmar, I., Hastings, J., & Martínez, P. (2015). DINTO: Using OWL Ontologies and SWRL Rules to Infer Drug-Drug Interactions and their Mechanisms. *Journal of Chemical Information and Modeling*, *55*(8), 1698-1707.

Hirankitti, V., & Xuan, T. M. (2011). A Meta-reasoning Approach for Reasoning with SWRL Ontologies. *Proc. of the International MultiConference of Engineers and Computer Scientists* (pp. 112-117, Vol I). Hong Kong: International Association of Engineers.

Hitzler, P., Krötzsch, M., Parsia, B., Patel-Schneider, P. F., & Rudolph, S. (2012). OWL 2 Web Ontology Language Primer (2nd Edition), W3C Rec, 11 Dec 2012. http://www.w3.org/TR/owl-primer

Horrocks, I., Parsia, B., Patel-Schneider, P. F., & Hendler, J. (2005). Semantic Web Architecture: Stack or Two Towers?. In *Proceedings of the 3rd International Workshop on Principles and Practice of Semantic Web Reasoning (PPSWR 2005)* (pp. 37-41). Dagstuhl, Germany:Springer, LNCS, Vol. 3703.

Horrocks, I., Patel-Schneider, P. F., Bechhofer, S., & Tsarkov, D. (2005). OWL rules: A proposal and prototype implementation. *Journal of Web Semantics*, *3*(1), 23-40.

Horrocks, I., Patel-Schneider, P. F., Boley, H., Tabet, S., Grosof, B., & Dean, M. (2004). SWRL: A Semantic Web Rule Language Combining OWL and RuleML. W3C Member Submission. 21 May 2004. http://www.w3.org/Submission/SWRL/

Jang, M., & Sohn, J.-C. (2004). Bossam: An Extended Rule Engine for OWL Inferencing. In *Proceedings of the International Workshop on Rules and Rule Markup Languages for the Semantic Web (RuleML 2004)* (pp. 128-138). Hiroshima, Japan: Springer, LNCS, Vol. 3323.

Khan, F., Bellandi, A., Frontini, F., & Monachini, M. (2017). Using SWRL Rules to Model Noun Behaviour in Italian. In *Proceedings of the First International Conference on Language, Data, and Knowledge (LDK 2017)* (pp. 134-142). Galway, Ireland: Springer, LNCS, Vol. 10318.

Kifer, M., & Boley, H. (2013). RIF Overview (2nd Ed), W3C Working Group Note, 5 Feb 2013. http://www.w3.org/TR/rif-overview/

Knublauch, H. (2011a). SPIN - Modeling Vocabulary, W3C Member Submission, 22 Feb 2011. http://www.w3.org/Submission/spin-modeling/

Knublauch, H. (2011b). SPIN - SPARQL Syntax, W3C Member Submission, 22 Feb 2011. http://www.w3.org/Submission/spin-sparql/

Knublauch, H., & Kontokostas, D. (Eds.). (2017). Shapes Constraint Language (SHACL), W3C Recommendation, 20 July 2017, https://www.w3.org/TR/shacl/.

Knublauch, H., Allemang, D., & Steyskal, S. (2017). SHACL Advanced Features, W3C Working Group Note 08 June 2017, https://www.w3.org/TR/shacl-af/.

Knublauch, H., Hendler, J. A., & Idehen, K. (2011). SPIN - Overview and Motivation, W3C Member Submission, 22 Feb 2011. http://www.w3.org/Submission/spin-overview/

Lloyd, J. W. (1987). *Foundations of logic programming* (2nd edition). Springer series in symbolic computation. Springer.

Matheus, C. J., Baclawski, K., Kokar, M. M., & Letkowski, J. (2005). Using SWRL and OWL to Capture Domain Knowledge for a Situation Awareness Application Applied to a Supply Logistics Scenario. In *Proceedings of the First International Conference on Rules and Rule Markup Languages for the Semantic Web (RuleML 2005)* (pp. 130-144). Galway, Ireland: Springer, LNCS, Vol. 3791.

Meditskos, G., & Bassiliades, N. (2008). A rule-based object-oriented OWL reasoner. *IEEE Transactions on Knowledge and Data Engineering*, *20*(3), 397–410.

Milanovic, M., Djuric, D., Gasevic, D., & Devedzic, V. (2009). Sharing Ontologies and Rules Using Model Transformations. *Handbook of Research on Emerging Rule-Based Languages and Technologies: Open Solutions and Approaches, Chapter 20* (pp. 471-492). IGI Global.

Motik, B., (n.d.), *KAON2*. http://kaon2.semanticweb.org/

Motik, B., Cuenca Grau, B., Horrocks, I., Wu, Z., Fokoue, A., & Lutz, C. (2012). OWL 2 Web Ontology Language Profiles (2nd Edition). W3C Recommendation, 11 Dec 2012. https://www.w3.org/TR/owl2-profiles/

Motik, B., Sattler, U., & Studer, R. (2005). Query Answering for OWL-DL with Rules. *Journal of Web Semantics*, *3*(1), 41-60.

Namahoot, C. S., Sivilai, S., Brückner, M. (2016). An Ingredient Selection System for Patients Using SWRL Rules Optimization and Food Ontology. In *Proceedings of the 13th International Conference on Cooperative Design, Visualization and Engineering* (*CDVE 2016)* (pp. 163-171). Sydney, Australia: Springer, LNCS, Vol. 9929.

Nools. (n.d.). https://github.com/C2FO/nools

O'Connor, M. J., Knublauch, H., Tu, S. W., Grossof, B., Dean, M., Grosso, W. E., & Musen, M. A. (2005). Supporting Rule System Interoperability on the Semantic Web with SWRL. In *Proceedings of the 4th International Semantic Web Conference (ISWC 2005)* (pp. 974-986). Galway, Ireland: Springer, LNCS, Vol. 3729.

O'Connor, M. J., Shankar, R. D., Nyulas, C., Tu, S. W., & Das, A. K. (2008). Developing a Web-Based Application using OWL and SWRL. In *Proceedings of the AAAI Spring Symposium: AI Meets Business Rules and Process Management* (pp. 93-98). Stanford University, California: AAAI Press.

Protégé ontology editor. (n.d.). http://protege.stanford.edu/.

RDFQuery. (n.d.). https://code.google.com/p/rdfquery/wiki/RdfPlugin

RDFStore-JS. (n.d.). https://github.com/antoniogarrote/rdfstore-js

REWERSE I1 Working Group. (2006). R2ML - The REWERSE I1 Rule Markup Language, https://oxygen.informatik.tu-cottbus.de/rewerse-i1/?q=node/6

Riazanov, A., & Voronkov, A. (2002). The design and implementation of VAMPIRE. *AI Communications*, *15*(2-3), 91-110.

Rigas, E., Meditskos, G., & Bassiliades, N. (2012). SWRL2COOL: Object-Oriented Transformation of SWRL in the CLIPS Production Rule Engine. In *Proceedings of the 7th Hellenic Conference on AI (SETN 2012)* (pp. 49-56). Lamia, Greece: Springer, LNCS, Vol. 7297.

RuleML. (n.d.). http://wiki.ruleml.org/index.php/RuleML_Home

Samavi, R., & Consens, M. P. (2018). Publishing privacy logs to facilitate transparency and accountability. *Journal of Web Semantics*, *50*, 1-20.

Samuel, K., Obrst, L., Stoutenberg, S., Fox, K., Franklin, P., Johnson, A., Laskey, K., Nichols, D., Lopez, S., & Peterson J. (2008). Translating OWL and semantic web rules into Prolog: Moving toward description logic programs. *Theory Practice of Logic Programming*, *8*(3), 301-322.

Sirin, E., Parsia, B., Grau, B. C., Kalyanpur, A., Katz, Y. (2007). Pellet: A Practical OWL-DL Reasoner. *Journal of Web Semantics*, *5*(2), 51-53.

Somodevilla, M. J., Mena, I., Pineda Torres, I. H., de Célis Herrero, C. P. (2015). Deducting Lifestyle Patterns by Ontologies' SWRL Rules. In *Proceedings of the Workshops of the 26th International Conference on Database and Expert Systems Applications* (*DEXA Workshops 2015)* (pp. 9-13). Valencia, Spain: IEEE.

Stardog Union. (2018). Stardog, https://www.stardog.com/

SWRL Drools Tab. (2012). http://protege.cim3.net/cgi-bin/wiki.pl?SWRLDroolsTab.

ter Horst, H. J. (2005). Completeness, decidability and complexity of entailment for RDF Schema and a semantic extension involving the OWL vocabulary. *Journal of Web Semantics*, *3*(2–3), 79-115,

TopQuadrant. (n.d.). *TopBraid Composer*, https://www.topquadrant.com/tools/IDE-topbraid-composer-maestro-edition/

Tsarkov, D., & Horrocks, I. (2006). Fact++ description logic reasoner: System description. In *Proceedings of the 3rd International Joint Conference on Automated Reasoning (IJCAR 2006)* (pp. 292-297). Seattle, WA, USA: Springer, LNCS, Vol. 4130.

van Woensel, W., al Haider, N., Ahmad, A. M., & Abidi S. S. R. (2014). A Cross-Platform Benchmark Framework for Mobile Semantic Web Reasoning Engines. In *Proceedings of the 13th International Semantic Web Conference (ISWC 2014)* (pp. 389-408). Trentino, Italy: Springer, LNCS, Vol. 8796.

W3C. (2013). *The Semantic Web Activity*. http://www.w3.org/2001/sw/.

Wang, X., Ma, Z.M., Zhang, F., & Yan, L. (2010). RIF Centered Rule Interchange in the Semantic Web. In *Proceedings of the 21st International Conference on Database and Expert Systems Applications (DEXA 2010)*. (pp 478-486). Bilbao, Spain: Springer, LNCS, Vol. 6261.

Wielemaker, J., Schrijvers, T., Triska, M., & Lager, T. (2012). SWI-Prolog. *Theory and Practice of Logic Programming*, *12*(1-2), 67-96.